\documentclass[letterpaper, 10 pt, journal, twoside]{IEEEtran}  

\usepackage{hyperref}

\usepackage{booktabs,multirow}
\usepackage{algorithm}
\usepackage[noend]{algpseudocode}

\usepackage[english]{babel}
\usepackage[pdftex]{graphicx}
\usepackage{adjustbox}
\DeclareGraphicsExtensions{.pdf,.jpg,.jpeg,.png}

\usepackage{color}
\usepackage{bm}
\usepackage[caption=false]{subfig}
\usepackage{amsmath} 
\usepackage{amssymb}  
\usepackage{physics}
\usepackage[table]{xcolor}
\usepackage{algcompatible}

\usepackage{tikz}
\usetikzlibrary{positioning}
\usetikzlibrary{calc}

\newcommand{\trsp}{{\scriptscriptstyle\top}}

\newcommand\copyrighttext{%
  \footnotesize \textcopyright 2021 IEEE. Personal use of this material is permitted.
  Permission from IEEE must be obtained for all other uses, in any current or future
  media, including reprinting/republishing this material for advertising or promotional
  purposes, creating new collective works, for resale or redistribution to servers or
  lists, or reuse of any copyrighted component of this work in other works.}

\newcommand\copyrightnotice{%
\begin{tikzpicture}[remember picture,overlay]
\node[anchor=south,yshift=10pt] at (current page.south) {\fbox{\parbox{\dimexpr\textwidth-\fboxsep-\fboxrule\relax}{\copyrighttext}}};
\end{tikzpicture}%
}

\begin{document}

\title{Learning Constrained Distributions of
Robot Configurations with Generative Adversarial Network}
\author{Teguh Santoso Lembono, Emmanuel Pignat, Julius Jankowski,  and Sylvain Calinon
\thanks{Manuscript received: October, 15, 2020; Revised January, 18, 2021; Accepted February, 13, 2021.}
\thanks{This paper was recommended for publication by Editor Nancy Amato upon evaluation of the Associate Editor and Reviewers' comments. This work was supported by the European Commission's Horizon 2020 Programme (MEMMO project, http://www.memmo-project.eu/, grant 780684).}
\thanks{The authors are with Idiap Research Institute, Martigny, Switzerland and with EPFL, Lausanne, Switzerland \texttt{\{name.surname\}@idiap.ch}}
\thanks{Digital Object Identifier (DOI): see top of this page.}
}

\markboth{IEEE Robotics and Automation Letters. Preprint Version. Accepted February, 2021}
{Lembono \MakeLowercase{\textit{et al.}}: Learning Constrained Distributions of
Robot Configurations with Generative Adversarial Networks}

\maketitle

\copyrightnotice

\begin{abstract}
\label{sec:abstract}
In high dimensional robotic system, the manifold of the valid configuration space often has a complex shape, especially under constraints such as end-effector orientation or static stability. We propose a generative adversarial network approach to learn the distribution of valid robot configurations under such constraints. It can generate configurations that are close to the constraint manifold. We present two applications of this method. First, by learning the conditional distribution with respect to the desired end-effector position, we can do fast inverse kinematics even for very high degrees of freedom (DoF) systems. Then, we use it to generate samples in sampling-based constrained motion planning algorithms to reduce the necessary projection steps, speeding up the computation. We validate the approach in simulation using the 7-DoF Panda manipulator and the 28-DoF humanoid robot Talos.
\end{abstract}


\section{Introduction}
\label{sec:introduction}

\IEEEPARstart{G}{enerative} Adversarial Network (GAN)~\cite{Goodfellow2016} is a powerful method to learn complex distributions. It is particularly popular in computer vision to learn the distribution of images from a dataset. Some of the applications include generating high resolution images~\cite{Ledig2017}, text-to-image synthesis~\cite{Reed2016}, and interactive art~\cite{Zhu2016}. Considering the recent success of deep learning techniques in robotics, e.g.,~\cite{finn2016deep}, we propose to adapt GAN to the context of robotics, i.e., to learn the distribution of valid robot configurations in constraint manifolds.

In robotics, \emph{configuration space} refers to the space of possible robot configurations that may include joint angles for revolute joints, joint translations for prismatic joints, and the base pose for floating-based robots. This concept is very important in motion planning, because the planning often needs to be done in the configuration space. Due to the presence of constraints, the valid configurations lie in some manifold of the configuration space, the shape of which can be complicated and often cannot be parameterized explicitly. Inequality constraints, such as obstacle avoidance, result in a manifold with non-zero volume, and the standard approach to sample from this manifold is to do rejection sampling. For equality constraints (such as fixed feet locations or end-effector orientation), however, the manifold volume is reduced to zero, as the manifold has lower dimension than the original space. Rejection sampling does not work in this case because there is zero probability that the random sample will be on the manifold. A common approach is to project the random samples to the manifold, and this projection step takes a significant portion of the planning computation time. By using GAN to learn the distribution of valid robot configurations on a manifold, we can sample from this manifold effectively, such that the generated samples are close to the desired manifold.

Having learned the valid distributions, we show that it can be used in two applications: inverse kinematics (IK) and sampling-based constrained motion planning. Analytical IK is typically only available for robots with 6-DoF or lower. For higher DoF robots, the most common method is to use numerical IK where we start with an initial robot configuration (often selected randomly by uniform sampling), and rely on optimization to find the configuration that reach the desired pose while satisfying the constraints. In our proposed framework, we show that GAN can produce good initial configurations that are close to achieving the target, resulting in a faster numerical IK with higher success rate as compared to uniform sampling initialization. Furthermore, sampling-based constrained motion planning also involves a high number of projections of random samples to the constraint manifold. We show that by replacing the uniform sampling with GAN, the planning time can be reduced significantly.

\begin{figure}[t!]
\centering
\subfloat[][]{\includegraphics[width=0.33\columnwidth]{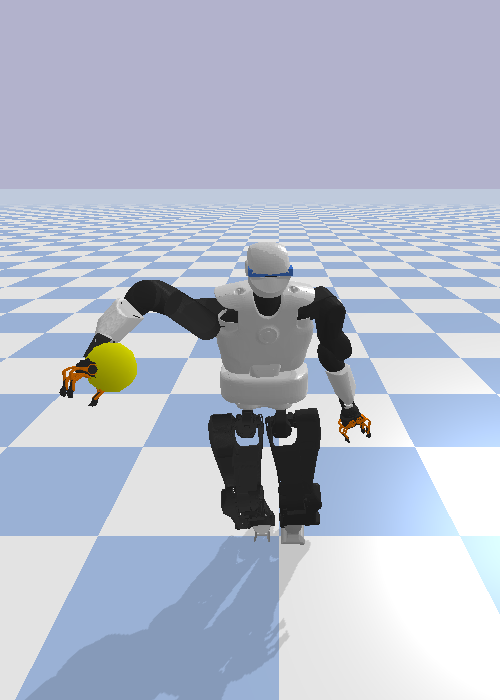}\label{fig:target_close}}
\hfill
\subfloat[][]{\includegraphics[width=0.33\columnwidth]{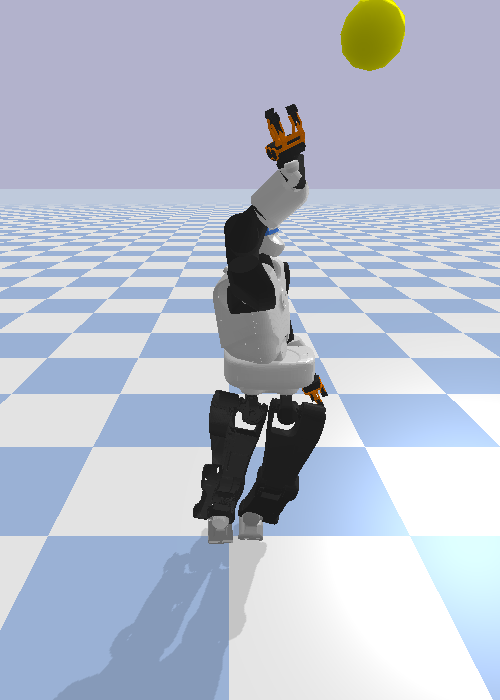}\label{fig:target_far}}
\hfill
\subfloat[][]{\includegraphics[width=0.33\columnwidth]{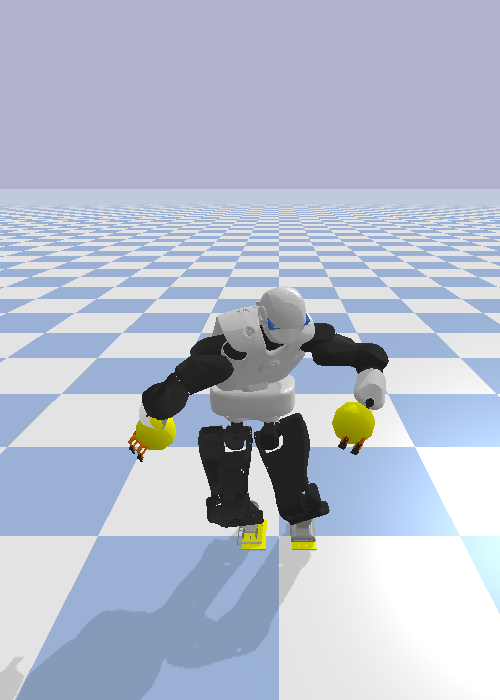}\label{fig:multilimbs}}
\caption{
Using GAN for obtaining approximate IK solutions. The targets are depicted in yellow. In \protect\subref{fig:target_close}, the target is reachable. When the target is out of reach \protect\subref{fig:target_far}, GAN still outputs a configuration close to the constraint manifold. We can also give the four limbs position as the IK targets \protect\subref{fig:multilimbs}. }
\label{fig:ik}
\end{figure}

\begin{figure*}
\label{fig:framework}
\centering
\includegraphics[scale=.8]{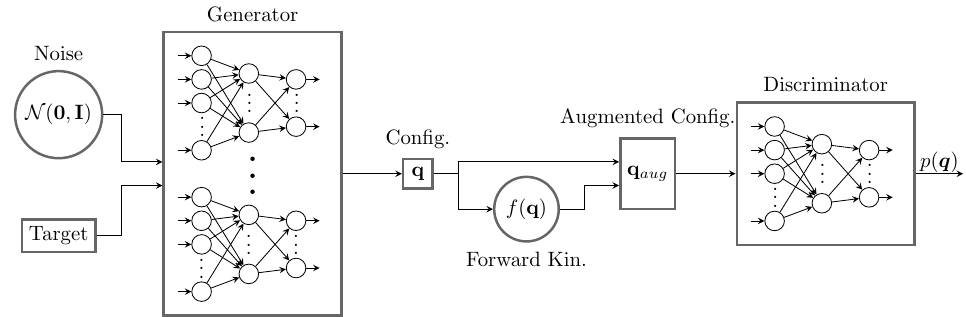}
\caption{The proposed GAN framework for learning the distribution of valid robot configurations. The generator consists of an ensemble of $N_\text{net}$ neural networks, while the discriminator consists of a single neural network. Besides a Gaussian noise as in standard GAN, we also add the end-effector target(s) as additional input to the generator. The output of the generator is then augmented by additional features, i.e., the corresponding end-effector poses, before being given to the discriminator.}
\end{figure*}

One particular difficulty of learning a distribution is when the target distribution is multimodal, as is the case in many robotic systems. For example, the conditional distribution of a 6-DoF manipulator given a desired end-effector pose is multimodal because there are many different configurations that are associated with the pose. To overcome this, we propose to use an ensemble of neural networks as the generator in GAN. Each neural network can converge to a different mode, and we get better coverage of the distribution as compared to using a single network.

The remainder of the paper is as follows. In Section~\ref{sec:related_work}, we review existing work on learning sampling distributions and constrained-based motion planning. Section~\ref{sec:method} describes the GAN framework and how we use it for inverse kinematics and constrained motion planning. The experiments with 7-DoF Panda and 28-DoF Talos~\cite{stasse2017talos} are presented in Section~\ref{sec:result}. Finally, we conclude with a few remarks in Section~\ref{sec:conclusion}.

\section{Related Work}
\label{sec:related_work}

\subsection{Learning sampling distribution}
In~\cite{Lehner2017,Lehner2018}, Gaussian Mixture Model (GMM) is used to learn the sampling distribution based on previously planned paths. The distribution is used to either generate biased samples for the sampling algorithm or to construct a repetition roadmap that guides towards finding the solution. It speeds up the computation as compared to uniform sampling, but it is difficult to generalize to different situations (e.g., different obstacles). Moreover, GMM does not scale well to higher dimensional systems. GMM is also used in~\cite{carpentier2017learning} to learn the feasibility constraint of center of mass (CoM) position with respect to the whole body dynamics. Conditional Variational Autoencoder (CVAE) is used in~\cite{Ichter2018}, also trained based on data from demonstrations or planned paths, but it is not implemented on constrained systems. Convolutional Autoencoder is used in~\cite{Holden2015} to learn the motion manifold of human motion, but not in the context of motion planning.

In contrast to the above approaches, we propose to use Generative Adversarial Network (GAN) to learn the sampling distribution of constrained robotic systems and apply it to the 7-DoF Panda and the 28-DoF Talos. GAN scales better with higher dimensions as compared to GMM, and it is easy to learn a conditional distribution with respect to some task (such as end-effector pose). We use it to initialize IK very efficiently while considering various constraints (joint limit, static stability, foothold location). A similar effort to initialize IK is done in \cite{tonneau2018efficient} by storing previously computed end-effector configurations in an octree data structure, indexed by the end-effector positions. However, in that work, each limb is treated separately from the body, and only the kinematics is considered without any stability criterion when retrieving the initial guess. Our GAN approach produces configurations that are already close to the constraint manifold, which can include the stability criterion. In~\cite{ren2020learning}, GAN is successfully used to learn inverse kinematics and dynamics of 8-DoF robot manipulator with the real data, but it does not consider any constraint on the robotic system. Furthermore, we show that the GAN sampler can also be used to improve the sampling-based constrained motion planning algorithm.

\subsection{Constrained motion planning}
A review of various approaches in sampling-based motion planning for constrained systems can be found in~\cite{Kingston2018}. Among the others, projecting the samples to the constraint manifold is a simple but very generalizable way of extending the standard Rapidly-exploring Random Tree (RRT) to constrained problems. Instead of doing rejection sampling, the samples are projected to the constraints manifold~\cite{Berenson2009, Berenson2011}. While it works well, the projection steps take most of the planning time. In~\cite{Yang}, Yang \emph{et al.} compare various algorithms for motion planning of humanoid robot. They report that the projection step takes more than 95\% of the planning time. Several research lines attempt to reduce this time computation. For example, in ~\cite{Stilman2010}, Stilman \emph{et al.} use the tangential direction of the constraint manifold to always move while staying close to the manifold. In~\cite{Kanehiro2012}, Kanehiro \emph{et al.} simplify the humanoid structure by splitting it into multiple 6-DoF structures and then perform analytical IK.

In our proposed method, we can generate samples that are already close to the constraints manifold, and the approach is generalizable to most robotic systems. This will reduce the necessary number of projection steps significantly and hence lower the computation time.
Additionally, optimization-based approaches such as CHOMP~\cite{Ratliff2009} and TrajOpt~\cite{Schulman2016} can solve constrained planning quickly by including the constraints in the optimization problem. However, since the problem is highly nonlinear, these methods often require good initial guesses, otherwise they may have a lower success rate even for simple problems~\cite{Lembono2020}. In contrast, sampling-based motion planning can find a global solution with probabilistic completeness guarantee~\cite{Berenson2011}, provided that the sampler can cover the entire feasible configuration space, which is the case for uniform sampling.

\section{Method}
\label{sec:method}
In this section, we present the proposed GAN framework for learning the robot distribution. We then propose two applications: inverse kinematics and constrained motion planning.

\subsection{Generative Adversarial Framework}
\label{sec:gan}
In the generative adversarial framework \cite{Goodfellow2016}, a generator $G(\bm{z}; \bm{\theta_G})$ is trained to transform the input noise $\{\bm{z}\}$ drawn from $p_z(\bm{z})$ (typically a unit Gaussian) into samples $\{\bm{q}\}$ that look similar to the data distribution. To do this, a discriminator $D(\bm{q}; \bm{\theta_D})$ is trained in parallel to output the probability $p(\bm{q})$ that tells whether $\bm{q}$ comes from the dataset or the generator. The training of GAN is therefore like a game between the generator and the discriminator where one tries to beat the other. The generator and discriminator are neural networks (with parameters $\bm{\theta_G}$ and $\bm{\theta_D}$, respectively) trained with Stochastic Gradient Descent (SGD).

In our approach, GAN is trained to generate configurations $\{\bm{q}\}$ that lie in some constraints manifold. The following sections explain some changes to standard GAN that we propose to better suit robotic applications. Unlike in images, we can more easily incorporate several forms of prior knowledge of what constitutes good configurations in the form of additional cost functions and transformations. To better handle multimodal distributions, we use an ensemble of neural networks as the generator. The framework is depicted in Fig.~\ref{fig:framework}.

\subsubsection{Additional inputs}
In standard GAN, the input to the generator is a noise sampled from a Gaussian distribution. To obtain a conditional distribution, we include the task parameters as additional inputs to the generator. The task parameters here correspond to the desired end-effector pose(s), but other additional tasks are also possible.

\subsubsection{Additional costs}
The training cost for the generator normally consists of the cost of tricking the discriminator to classify its samples as dataset samples. In robotics, however, we can add other costs that can be used to evaluate the quality of the samples based on the knowledge of the robotic system. Here we include several costs:
\begin{itemize}
\item The cost of end effector targets $c_\text{ee}(\bm{q})$. From the samples generated by $G$, we can compute the forward kinematics to obtain the end-effector positions and compare this to the desired end-effector target (given as the input to the generator).
\item The cost of static stability $c_\text{s}(\bm{q})$. To achieve static stability, the CoM projection on the ground should be located inside the foot support polygon.
\item The cost of joint limits $c_l(\bm{q})$. The cost is zero if it is within the limit, and increasing outside the limit.
\end{itemize}

\subsubsection{Output Augmentation}
Instead of feeding the configurations directly to the discriminator, we augment the configurations by some transformations, e.g., end-effector poses. This helps the discriminator to discern between good and bad samples according to the relevant features. Other transformations such as CoM location can also be added.

\subsubsection{Ensemble of networks}
When the desired distributions are multimodal, GAN often converges to only some modes of the distribution. This is known as the Helvetica scenario or mode collapse~\cite{Goodfellow2016}. For example, when the desired distribution is a GMM, GAN may converge to only some of the mixture components, and not all of them. This is a major problem in the robotics context, especially for motion planning, because omitting some portion of the configuration space means reducing the probability of finding feasible solutions. To overcome this, we use an ensemble of $N_\text{net}$ neural networks as the generator. Given an input, each network generates a corresponding robot configuration, and each one is trained as a stand-alone generator. When the desired distribution is multimodal, each network may converge to a different mode.

\begin{figure}[t!]
\centering
	\subfloat[][]{\includegraphics[width=0.33\columnwidth]{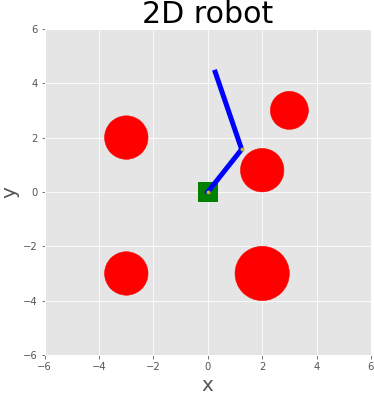}\label{fig:2drob}}
\hfill
\subfloat[][]{\includegraphics[width=0.33\columnwidth]{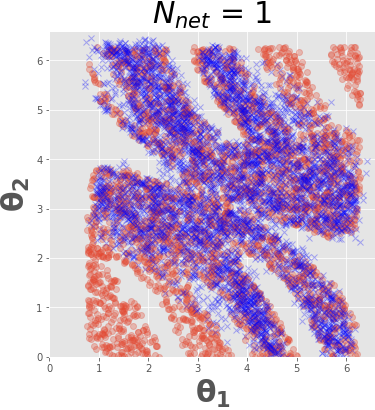}\label{fig:2drob1}}
\hfill
\subfloat[][]{\includegraphics[width=0.33\columnwidth]{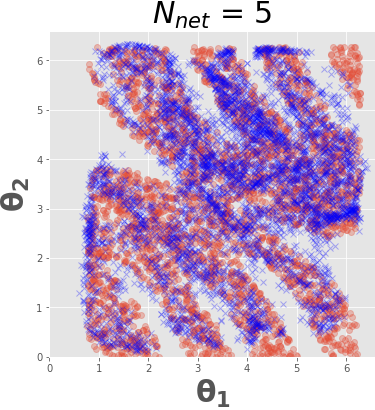}\label{fig:2drob5}}
\caption{ Illustrative example of a 2-DoF robot with obstacles. \protect\subref{fig:2drob} shows the robot with circular obstacles. The configurations (i.e.,\ joint angles) that are not in collision are plotted in \protect\subref{fig:2drob1} and \protect\subref{fig:2drob5} as red circles. We learn this distribution using GAN. In \protect\subref{fig:2drob1}, we use one neural network as the generator, and the GAN samples are plotted as blue crosses. We see that the samples do not cover the whole distribution. Using an ensemble of 5 networks in \protect\subref{fig:2drob5}, we manage to cover most of the distribution.}
    \label{fig:2drob_example}
\end{figure}

The advantage of using an ensemble of networks as the generator can clearly be seen using an illustrative 2-link robot example. Fig.~\ref{fig:2drob} shows the setup of the robot surrounded by obstacles. The configuration consists of the two joint angles. The valid configurations (i.e.,\ the ones without collision) are plotted in Fig.~\ref{fig:2drob1} and Fig.~\ref{fig:2drob5} as red circles, and GAN is trained to learn this distribution. When using only one network for the generator, GAN converges to only some part of the configuration space, as depicted in Fig.~\ref{fig:2drob1} (the GAN samples are plotted as blue crosses). Using $N_\text{net} = 5$ networks, we manage to cover most of the configuration space (Fig.~\ref{fig:2drob5}).

\subsection{Inverse Kinematics}
\label{sec:ik}
For high-dimensional robots where analytical IK is not tractable, numerical IK is the standard approach. Given a desired end-effector pose $\bm{p}_{\text{ref}}$ and the initial configuration $\bm{q}_0$, an iterative optimization is used to find the value of $\bm{q}$ such that the corresponding end effector pose $\bm{p}(\bm{q})$ is equal to the desired pose $\bm{p}_{\text{ref}}$. One common approach is to use the Gauss-Newton Algorithm, which we use in this paper.
At its most basic formulation, the inverse kinematics is formulated as minimizing the following quadratic cost,
\begin{equation}
    c(\bm{q}) = \frac{1}{2}\bm{r}^{\trsp}\bm{r},
\label{eq:cost_function}
\end{equation}
where $\bm{r} = \bm{p}(\bm{q}) - \bm{p}_{\text{ref}}$ is the residual vector that we want to reduce to zero. Starting from the initial guess $\bm{q}_0$, the next solution is obtained by
\begin{equation}
    \bm{q}_{i+1} = \bm{q}_{i}- \alpha\bm{J}^\dagger \bm{r},
\label{eq:opt_steps_std}
\end{equation}
where $\bm{J} = \pdv{\bm{r}}{\bm{q}}$ is the Jacobian of the residual function, $\alpha$ is the step length, and the subscript $^\dagger$ denotes the (Moore-Penrose) pseudo-inverse operator. The step is iterated until the residual norm is smaller than a specified threshold value.

This formulation can be extended to more residual functions, each of which corresponds to a particular constraint that we want to enforce to the system.
For example, in the case of IK for humanoid, we can define the cost function as
$$c(\bm{q}) = \frac{1}{2}(\bm{r}_\text{ee}^\trsp \bm{r}_\text{ee} + \bm{r}_\text{s}^\trsp \bm{r}_\text{s} + \bm{r}_\text{l}^\trsp \bm{r}_\text{l}),$$
where the subscripts ($\text{ee}, \text{s}, \text{l}$) refer to the residuals corresponding to the constraints on the end-effector pose, static stability, and joint limit. Note that the Gauss Newton algorithm can only be applied to least-square problems, hence each of the cost term should be a square function of the residual.

Furthermore, with the pseudo-inverse method, we can implement some hierarchy in the constraints by using the nullspace projection~\cite{escande2010fast}. For each level of the hierarchy, we can define a cost function as in~\eqref{eq:cost_function}. Let $\bm{J}_1$ and $\bm{J}_2$ refer to the Jacobian of the residuals of the main and secondary constraints. The nullspace projection operator due to the first Jacobian, $\bm{N}_1$, can be obtained as $\bm{N}_1 = \bm{I} - \bm{J}_1^\dagger \bm{J}_1$. We can use this nullspace operator to prevent the secondary constraints from affecting the main constraints.
Starting from the initial guess $\bm{q}_0$, the next solution is obtained by
\begin{equation}
    \bm{q}_{i+1} = \bm{q}_{i} - \alpha_1\bm{dq}_1 - \alpha_2\bm{dq}_2,
    \label{eq:opt_step_hierarchy}
\end{equation}
where $\bm{dq}_1$ and $\bm{dq}_2$ are the steps corresponding to the main and secondary constraints, defined as
$$ \bm{dq}_1 = \bm{J}_1^\dagger \bm{r}_1, \quad \text{and} $$
$$ \bm{dq}_2 = (\bm{J}_2 \bm{N}_1)^\dagger (\bm{r}_2 - \alpha_1\bm{J}_2 \bm{dq}_1),$$
and $(\alpha_1, \alpha_2)$ are the corresponding step lengths. In this work, we use the secondary constraint to maintain the configuration around a nominal posture $\bm{q}_{\text{nom}}.$ More details about the cost functions can be found in the appendix.

To determine the step length $\alpha_1$ and $\alpha_2$, we need to perform line search. We use the Armijo condition~\cite{nocedal2006numerical} as the stopping criteria. Finally, to improve the stability of the pseudo-inverse, we add a damping term $\lambda\bm{I}$, defined as
\begin{equation}
    \lambda = \mu \bm{r}^\trsp \bm{r} + \bar{\mu},
\label{eq:damping_term}
\end{equation}
where $\mu$ and $\bar{\mu}$ are manually-defined constants. The damping term hence depends on the residual magnitude. As shown in~\cite{sugihara2011solvability} and also observed in our experiments, it helps the convergence when starting far from the optimum solution.

The computation time for numerical IK really depends on how close the initial guess is to the optimal solution. As discussed in Section~\ref{sec:gan}, we can obtain good initial guesses by sampling from GAN while giving the end-effector poses as additional inputs to the generator. The resulting configurations would be close to the desired poses and the constraint manifold, reducing the required number of IK iterations significantly.

Finally, although the formulation here is presented for inverse kinematics, the same formulation can be used to project robot configurations to the constraint manifold by omitting the cost on the end-effector position. Further details will be provided in Section~\ref{sec:result}.

\subsection{Constrained Motion Planning}
\label{sec:crrt}
The standard approach in sampling-based constrained motion planning~\cite{Berenson2011} is largely based on RRT, with the addition of the projection step; each sample to be added to the tree is projected first to the constraint manifold. The projection step is formulated as an optimization problem as discussed in Section~\ref{sec:ik}. Algorithm~\ref{tab:crrt} describes the steps for constrained RRT (cRRT) for reaching a goal in task space, starting from an initial configuration $\bm{q}_0$. We refer to~\cite{Berenson2011} for more details of the algorithm.
 
First, we start with a tree $G$ initialized with the node $\bm{q}_0$. From the given goal task in Cartesian space, we compute $K$ goal configurations (by numerical IK). The following iterations then attempt to extend the tree to one of these goals. At each iteration, a random sample $\bm{q}_\text{rand}$ is generated. Nearest neighbor algorithm is used to find the nearest node $\bm{q}^a_\text{near}$ in the tree, and we then extend the tree from $\bm{q}^a_\text{near}$ to ${\bm{q}}_\text{rand}$. The last configuration obtained from the extension step is denoted as $\bm{q}^a_\text{reached}$. Next, we extend the tree from $\bm{q}^a_\text{reached}$ to one of the goal configurations, $\bm{q}_{g,k}$, chosen to be the one nearest to $\bm{q}^a_\text{reached}$. The last configuration obtained from the extension step is denoted as $\bm{q}^b_\text{reached}$. If $\bm{q}^b_\text{reached}$ is equal to $\bm{q}_{g,k}$, we stop the iteration, and compute the path from the root node $\bm{q}_0$ to $\bm{q}_{g,k}$. Otherwise, we continue with the next iteration until the goal is reached or the maximum number of iteration is exceeded.

In the extension step, we iteratively move from $\bm{q}^a_\text{near}$ to ${\bm{q}}_\text{rand}$ with a step size $\Delta q_{\text{step}}$, project the resulting configuration to the manifold, and check for collision. The extension stops when the projection fails or it is in collision.

\subsubsection{GAN sampling}
As reported in~\cite{Yang}, the projection steps dominate the computation time with more than 95\% of the total time. Instead of uniform sampling, we propose to use the GAN framework to generate the samples. This will give us samples that are already quite close to the manifold and hence reduce the computation time significantly.

To generate samples from the GAN framework, we first determine the task space region of interest, i.e.,\ a box that covers the reachable points of the robot's end-effector. We sample points inside this box and use it as the target for the generator. Together with the Gaussian noise, we can then obtain a set of configurations that are near to the target and require fewer projection steps. As the generator consists of $N_\text{net}$ neural networks, we choose one out of the $N_\text{net}$ configurations randomly as the output of the sampler.

\begin{algorithm}[t]
\small
\caption{Constrained RRT with goal sampling}\label{tab:crrt}
\begin{algorithmic}[0]
\State \textbf{INPUT}: $\bm{q}_0,\bm{x}_g$
\State \textbf{OUTPUT}: the path $\{\bm{q}_i\}|_{i=0}^{T}$
\end{algorithmic}

\begin{algorithmic}[1]
\State $G.init(\bm{q}_0)$\;
\State {$\{\bm{q}_{g,i}\}|_{i=0}^{K} \gets$} SampleGoal($\bm{x}_g$)
\WHILE{$n < $ max\_iter}
\State $\bm{q}_\text{rand} \gets$ SampleConfig()\;
\State $\bm{q}_\text{near} \gets$ NearestNeighbor($G, {\bm{q}}_\text{rand}$	)\;
\State $\bm{q}^a_\text{reached} \gets$ ConstrainedExtend($G, \bm{q}_\text{near}, {\bm{q}}_\text{rand}$	)\;
\State $\bm{q}_{g,k} \gets$ NearestNeighbor($\{\bm{q}_{g,i}\}|_{i=0}^{K}, \bm{q}^a_\text{reached}$	)\;
\State $\bm{q}^b_\text{reached} \gets$ ConstrainedExtend($G, \bm{q}^a_\text{reached}, \bm{q}_{g,k}$	)\;
\IF{$\bm{q}^b_\text{reached}$  = $\bm{q}_{g,k}$}
\State P $\gets$ ExtractPath($T, \bm{q}_0,\bm{q}_{g,k}$)
\State return P
\ENDIF
\ENDWHILE
\end{algorithmic}
\end{algorithm}

\section{Experimental Results}
\label{sec:result}

We implemented the proposed method on two systems: the 7-DoF Panda and 28-DoF Talos. The dataset is created by uniformly sampling random configurations and then projecting them to the constraints manifold. A data point corresponds to a robot configuration that satisfies the specified constraints. We use $N= 25000$ samples to train GAN for both Panda and Talos. The training time takes under 15 minutes, which is quite fast due to the additional cost functions in Section III.A that help the convergence of the GAN training. The generator consists of $N_\text{net} = 10$ neural networks with 2 hidden layers, each has 200 nodes, while the discriminator is a neural network with 2 hidden layers (each has 20 nodes for Panda and 40 nodes for Talos). $N_\text{net}$ is determined by training the generator several times with different numbers of neural networks and choose the one with the best performance on the motion planning task. ReLu is used as the activation function. We train the networks using SGD. All experiments\footnote{The implementation codes are available at \url{https://github.com/teguhSL/learning_distribution_gan}} are run on a processor Intel i7-8750H CPU $@$ 2.20GHz $\times$ 12.

\renewcommand{\arraystretch}{1.}
\begin{table*}[!t]
	\centering
      \caption{Comparing Projection and IK initialized by uniform sampling vs GAN sampling. The asterisk signifies that the corresponding values are computed only for the successful trials.}
      \label{tab:ik_result}
		
	\begin{tabular}{l l l | c  c  c  c }
		\toprule

\bf{Robot} & \bf{Task} & \bf{Sampling Method} & \textbf{Success } & $\bm{T}_\text{ave} (ms)$ & $\bm{T}^*_\text{ave} (ms)$ & $\textbf{Opt. Steps}^*$ \\
       \midrule

\multirow{4}{*}{Panda} &       \multirow{2}{*}{Projection} & Uniform & 98.6 &  4.2 $\pm$ 6.8 &  3.6 $\pm$ 3.6  &  6.5 $\pm$ 6.2\\
& & GAN & 100.0 &  1.0 $\pm$ 0.2 &  1.0 $\pm$ 0.2 &  1.9 $\pm$ 0.4\\
		
    &   \multirow{2}{*}{IK}  & Random & 76.4 &  29.3 $\pm$ 43.5 &  8.8 $\pm$ 7.5 &  7.9 $\pm$ 5.4\\
& & GAN & 88.6 &  12.5 $\pm$ 30.2 &  2.2 $\pm$ 2.2 &  2.9 $\pm$ 1.7\\
\midrule

\multirow{4}{*}{Talos} &       \multirow{2}{*}{Projection} & Uniform & 84.4 &  20.5 $\pm$ 25.3 &  10.5 $\pm$ 9.8 &  8.7 $\pm$ 7.3\\
& &  GAN & 100.0 &  2.1 $\pm$ 1.2 &  2.1 $\pm$ 1.2 &  2.1 $\pm$ 1.0\\
		
    &   \multirow{2}{*}{IK}  & Uniform & 82.6 &  28.7 $\pm$ 31.1 &  15.7 $\pm$ 13.0 &  10.4 $\pm$ 7.8\\
& &  GAN & 100.0 &  2.8 $\pm$ 0.7 &  2.8 $\pm$ 0.7 &  2.5 $\pm$ 0.6\\
		\bottomrule

	\end{tabular}
\end{table*}

\renewcommand{\arraystretch}{0.8}
\begin{table*}[!t]
	\centering    
      \caption{Comparing constrained RRT using uniform sampling vs GAN sampling.}
      \label{tab:crrt_result}
		
	\begin{tabular}{ l l  c  c   c  c c}
		\toprule
		\centering		
		{\bf{Robot}} & {\bf{Sampling Method}} & $\textbf{Success}$ & $\bm{T}_\text{ave} (s)$ & $\textbf{Opt. Steps}$ & $\bm{E}$\\
       \midrule
		\centering
	 \multirow{2}{*}{Panda }	& Random & 100.0 & 1.44 $\pm$ 1.23 & 2065.7 $\pm$ 1763.2 & 116.5 $\pm$ 93.6 \\
& GAN & 99.0 & 0.74 $\pm$ 0.66 & 902.5 $\pm$ 796.5 & 59.7 $\pm$ 60.7  \\
& Hybrid & 100.0 & 0.90 $\pm$ 0.77 & 1200.3 $\pm$ 1036.8 & 68.1 $\pm$ 56.6  \\
       \midrule
		\centering
      \multirow{2}{*}{Talos (Task 1)} & Uniform & 100.0 & 1.20 $\pm$ 0.99 & 464.4 $\pm$ 390.7 & 16.2 $\pm$ 12.4 \\
& GAN & 100.0 & 0.28 $\pm$ 0.13 & 74.0 $\pm$ 34.3 & 10.2 $\pm$ 7.3 \\
& Hybrid & 100.0 & 0.43 $\pm$ 0.25 & 131.4 $\pm$ 80.0 & 13.5 $\pm$ 10.0 \\
       \midrule
		\centering		
	   \multirow{3}{*}{Talos (Task 2)} & Uniform & 100.0 & 0.92 $\pm$ 0.82 & 327.8 $\pm$ 306.7 & 13.9 $\pm$ 13.1 \\
& GAN & 100.0 & 0.60 $\pm$ 0.35 & 127.0 $\pm$ 74.9 & 36.7 $\pm$ 29.4 \\
& Hybrid & 100.0 & 0.66 $\pm$ 0.44 & 182.3 $\pm$ 130.0 & 25.9 $\pm$ 19.5 \\

       \midrule
	   \multirow{2}{*}{Talos (Task 3)} 
& Uniform & 100.0 & 3.94 $\pm$ 3.63 & 1154.0 $\pm$ 1083.2 & 98.7 $\pm$ 72.4 \\
& GAN & 100.0 & 1.05 $\pm$ 1.43 & 179.0 $\pm$ 197.0 & 52.1 $\pm$ 158.6 \\
& Hybrid & 100.0 & 1.11 $\pm$ 0.94 & 228.7 $\pm$ 203.0 & 40.8 $\pm$ 51.1 \\		\bottomrule
	\end{tabular}
\end{table*}

\subsection{Projection and Inverse Kinematics (IK)}
\label{sec:ik_result}
As described in Section~\ref{sec:ik}, the projection and IK are formulated as optimization problems by defining several cost functions based on the desired constraints. The optimization problem is solved using Gauss-Newton algorithm. $\mu$ and $\bar{\mu}$ are set to $10^{-4}$ and $10^{-6}$.  We compare initializing the projection and IK by GAN sampling against uniformly sampling random configurations within the joint limits, which we denote as \emph{uniform} sampling. We set a certain threshold for each cost function, and the optimization is run until all the residuals are below the thresholds.

\paragraph{Panda}
Panda's configuration consists of 7 joint angles. The main cost function for IK consists of 3 terms: a) joint limit, b) EE orientation constraint, and c) EE position. The orientation is constrained such that the gripper is always in the horizontal position. We also add a secondary cost function, i.e., a posture cost that regularizes around a nominal configuration. The projection has the same set of cost functions as IK, except for the EE position.

\paragraph{Talos} 
Talos' configuration consists of 28 joint angles (7 for each arm, 6 for each leg, and 2 for the torso) and 6 numbers for the base pose. The main cost function for IK consists of the following terms: a) joint limit, b) static stability, c) the feet pose, and d) the right-hand position. The secondary cost function is defined as the posture cost around the nominal configuration, which is chosen to be the initial configuration given to the IK solver except for the left arm (which is regularized around a default posture). The EE is set to be at the right hand. The feet are constrained to remain at the same location. The task is to reach the desired location of the right hand, while respecting the constraints. For the projection, we omit the cost on the EE position.

We evaluate the methods with $N = 500$ tasks, and the result is shown in Table~\ref{tab:ik_result}. $\bm{T}_\text{ave}$ is the average computation time when considering all tasks, while $\bm{T}^*_\text{ave}$ and $\text{Opt. Steps}^*$ denote the average computation time and the number of optimization steps of only the successful tasks. We can see that using GAN speeds up both projection and IK computation significantly, around 2-5 times faster than uniform sampling, even when considering only the successful results. GAN samples only require around 2-4 optimization steps to achieve convergence. Uniform sampling also has a lower success rate, as can be expected for a nonlinear optimization problem (starting far from the optimal solution reduces the success rate). When the optimization cannot find the solution, it continues optimizing until reaching the maximum iteration, hence spending high computational time (namely, $\bm{T}_\text{ave}$ is higher than $\bm{T}^*_\text{ave}$). In practice, $\bm{T}_\text{ave}$ is the one we actually observe, since there is no way to avoid bad random samples.

\begin{figure}[t!]
\centering
\subfloat[][]{\includegraphics[height=0.2\columnwidth]{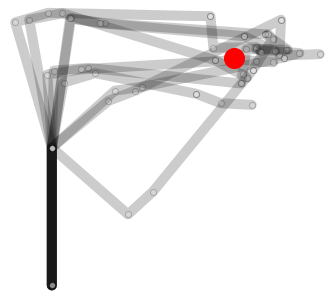}\label{fig:panda_config}}
\hspace{1cm}
\subfloat[][]{\includegraphics[height=0.3\columnwidth]{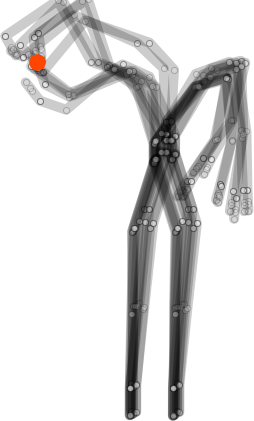}\label{fig:talos_config}}
 \\
\subfloat[][]{\includegraphics[height=0.2\columnwidth]{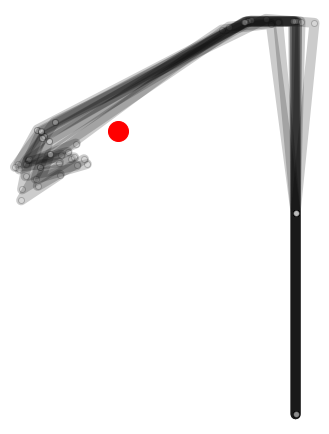}\label{fig:panda_config1}}
\hspace{1.5cm}
\subfloat[][]{\includegraphics[height=0.27\columnwidth]{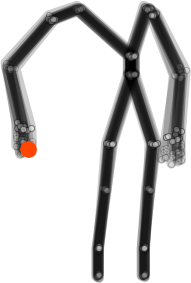}\label{fig:talos_config2}}
\caption{
Samples generated by GAN for Panda~\protect\subref{fig:panda_config} and Talos ~\protect\subref{fig:talos_config} using the proposed GAN framework. The samples correspond to the desired end-effector positions as shown in red. GAN manages to generate multimodal configurations with large variance. In contrast,~\protect\subref{fig:panda_config1} and~\protect\subref{fig:talos_config2} shows the samples generated by the same framework but when the discriminator is omitted. We see here that the ensemble of networks converges to only one mode with very low variance both for Panda \protect\subref{fig:panda_config1} and Talos~\protect\subref{fig:talos_config2}.}
\label{fig:ensemble}
\end{figure}

\begin{figure*}[t!]
\centering
\subfloat[][]{\includegraphics[height=0.45\columnwidth]{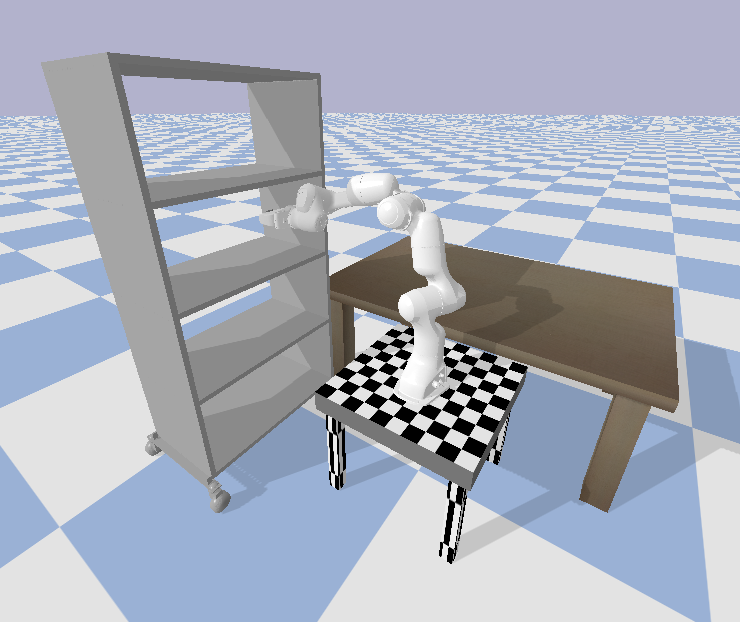}\label{fig:panda_task1}}
\hspace{0.5cm}
\subfloat[][]{\includegraphics[height=0.45\columnwidth]{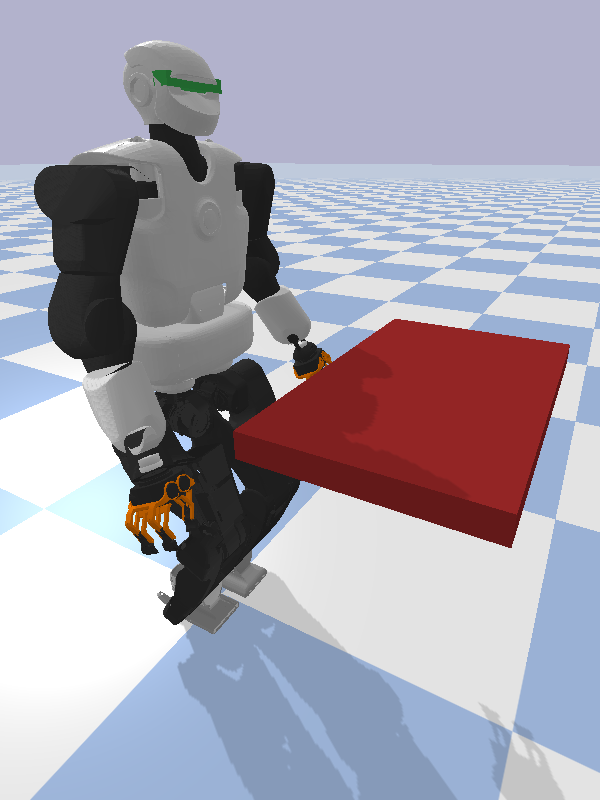}\label{fig:task1}}
\hspace{0.5cm}
\subfloat[][]{\includegraphics[height=0.45\columnwidth]{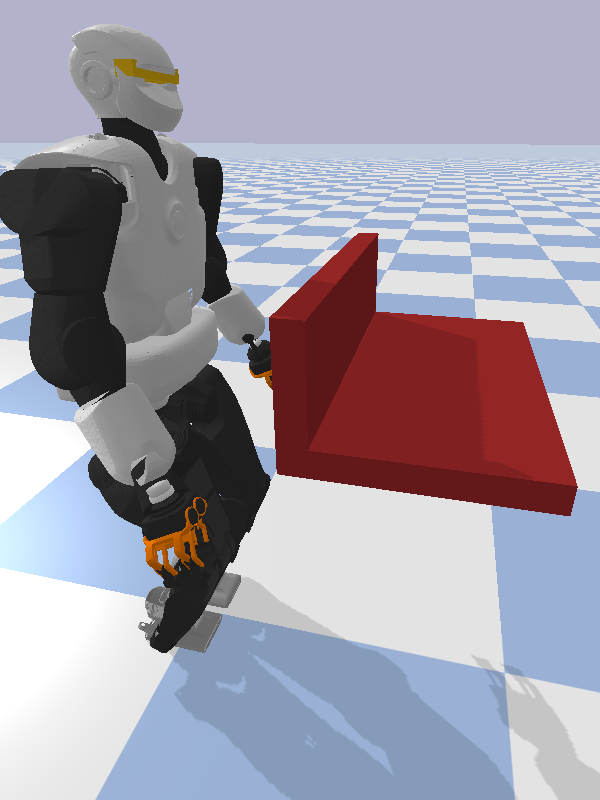}\label{fig:task2}}
\hspace{0.5cm}
\subfloat[][]{\includegraphics[height=0.45\columnwidth]{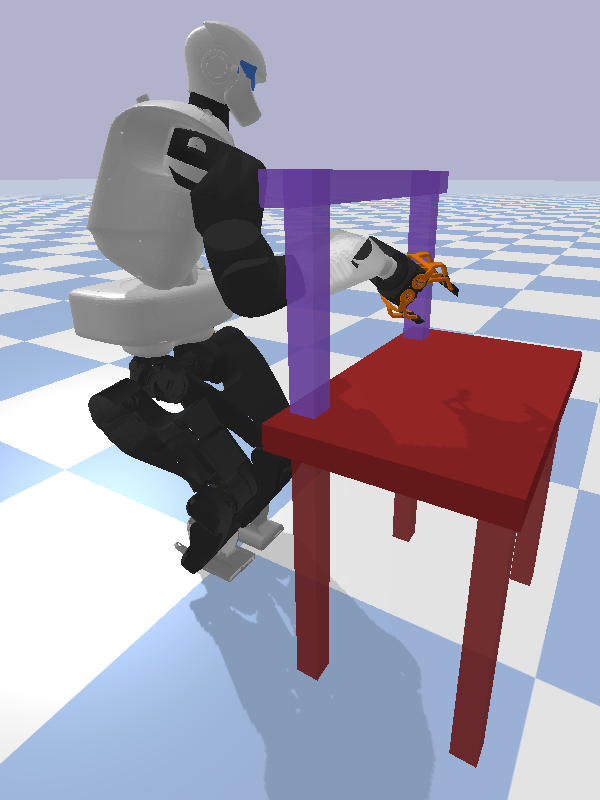}\label{fig:task3}}
\caption{ Constrained motion planning tasks for Panda \protect\subref{fig:panda_task1} and Talos (\protect\subref{fig:task1}, \protect\subref{fig:task2}~and \protect\subref{fig:task3}).}
\label{fig:crrt}
\end{figure*}

Besides the quantitative results shown in Table~\ref{tab:ik_result} and \ref{tab:crrt_result}, we observe that GAN produces configurations that are still close to the manifold even when the target is infeasible, i.e., too far from the arm reach (Fig.~\ref{fig:target_far}). Additionally, we can also extend the target variables to include the left hand and both feet, so that we can do approximate IK for all the four limbs simultaneously using GAN (Fig.~\ref{fig:multilimbs}). We refer to the supplementary video for better visualization of the infeasible target IK and the multi-limbs IK.

\subsection{Motion Planning}
\label{sec:crrt_result}
To use the GAN sampling in a sampling-based motion planner, the sampler must have good coverage of the distribution, especially when it is multimodal. We have shown in Fig.~\ref{fig:2drob_example} that using the ensemble of neural networks help GAN to cover a complex distribution. In addition to this, the discriminator also helps to increase the coverage. Indeed, without providing the dataset to the discriminator and train it together with the generator, even the ensemble of neural networks cannot have good coverage of the distributions, and often they converge to one mode only.
Fig.~\ref{fig:ensemble}a-b shows samples generated by GAN for Panda and Talos by giving the desired EE position (shown in red). We can see that in both robots the generated samples belong to multiple modes with good variance. In contrast, the samples in Fig.~\ref{fig:ensemble}c-d shows samples generated by GAN while removing the discriminator and training using only the additional cost functions in Section III.A (to be strict in terminology, this makes it no longer an adversarial network, but this is done to demonstrate the necessity of the discriminator). Without the discriminator, the samples do not have large variance and they converge to only one mode, despite using the ensemble of networks.

We then use GAN sampling in sampling-based constrained motion planning for both Panda and Talos, as described in Algorithm~\ref{tab:crrt}. For Panda, the task is to move the end-effector from above the table to one of the shelves, as shown in Fig.~\ref{fig:crrt}a, while maintaining the gripper in the horizontal position. For Talos, we consider three different environments in Fig.~\ref{fig:crrt}b-d in increasingly more difficult order, similar to the ones used in~\cite{Yang}. The task is to move from a random initial configuration above a table to reach a specified target position with the right hand below the table while satisfying the static stability and joint limits. For each environment, we run $N=100$ random tasks. We compute $K=10$ goal configuration for each task. Each task is run until it is successful or it reaches the maximum number of extension steps, which is set to be 500. In the case of failure, the planning is repeated until a maximum of two times and the total time is taken.

We compare three different sampling methods: \emph{uniform}, \emph{GAN}, and \emph{hybrid} sampling. Hybrid sampling is a combination of uniform and GAN sampling. It outputs GAN samples with a probability of $p$ and uniform samples with a probability of $1-p$. By including the uniform sampling, we can keep the probabilistic completeness guarantee of the planner. From the experiment, we observe that a value of $p = 0.8$ performs the best. An adaptive value is also possible, i.e., starting with $p = 1$ and decreasing it as the number of extensions grows, such that it converges to uniform sampling when the planner still cannot find any solution after a long time.

The result can be seen in Table~\ref{tab:crrt_result}. $\bm{T}_\text{ave}$, $\text{Opt. Steps}$, and $\bm{E}$ denote the average planning time, the number of optimization steps, and the number of extension steps in planning (\emph{ConstrainedExtend} in Algorithm~\ref{tab:crrt}) for one task.

For both Panda and Talos, GAN results in a significant reduction in the computation time, around 2-4 times faster than the uniform sampling. This gain is mainly due to the fewer optimization steps necessary to project the resulting configurations, as discussed in the previous section. The high success rate of GAN sampling also indicates that it manages to cover a good proportion of the configuration space, at least for the tasks that we consider here.

On the other hand, the comparison between GAN and hybrid sampling strategy is quite interesting. In most cases, GAN is still faster than hybrid sampling. However, if we look at the number of extension steps, hybrid sampling sometimes requires fewer extension steps than GAN. We also observe that GAN sometimes fails a task, while hybrid sampling is successful, such as the case in the Panda experiment. This shows that hybrid sampling can explore the distribution more effectively than GAN in these tasks. Its overall computation time is still higher than GAN, though,  because it requires more optimization steps due to the uniform sampling components.

\subsection{Discussion}
\label{sec:discussion}

From the experiment results, we show that GAN can learn to generate good quality samples close to the desired manifold. In addition, we can conclude from the motion planning results that it has good coverage over the distribution of robot configurations. There is no guarantee, however, that the distribution is perfectly covered, so in some rare cases the motion planning may fail to find a feasible solution, but using a hybrid sampling strategy recovers the completeness guarantee.

We manage to get good performance even with quite a basic GAN structure. Note that the GAN framework is trained without considering collision, unlike in the example of the 2-link robot in Section III.A. This means that the resulting sampler does not depend on the environment, and it works directly in any environment even with moving obstacles. We also show that the framework is easily adapted to different robots by the experiments on the Panda and Talos robot. Given any new robot and its constraints, we only need to formulate the cost functions corresponding to the constraints and generate the dataset, then the GAN framework can be trained quickly.

The stability criteria that we use here is static stability. In~\cite{ren2020learning} and~\cite{Pignat20CORL}, GAN has been shown to work well for problems with dynamics, so it would potentially be possible to extend our approach to generate more dynamical motions by including dynamic stability criteria such as Zero Moment Point (ZMP)~\cite{kajita2003biped} or Contact Wrench Cone Margin~\cite{dai2016planning}.

Since GAN is a very flexible tool, there are a lot of potential improvements. As in~\cite{Ichter2018}, by generating many planning data in a given environment, we can condition GAN on initial and goal configurations to obtain samples that are relevant for the task, instead of trying to cover the whole distributions. GAN also works well with high dimensional inputs such as images, so it would be possible to train with the environment representation (e.g.,\ voxel data or heightmap) to generate samples that avoid collisions in different environments. Since GAN can be conditioned on the desired target location, we can also use it to generate biased samples in task space.

In this paper, we use an ensemble of networks as the generator to cover the multimodal distribution. There are also other methods that attempt to handle this mode collapse issue, e.g.,\ using Wasserstein loss~\cite{arjovsky2017wasserstein} or unrolled GANs~\cite{Liu2016}. We will investigate these approaches in our future work.

\section{Conclusion}
\label{sec:conclusion}
We have presented a GAN framework to learn the distribution of valid robot configurations under various constraints. The method is then used for inverse kinematics and sampling-based constrained motion planning to speed up computational time. We validate the proposed method on two simulated platforms: 7-DoF manipulator Panda and 28-DoF humanoid robot Talos. We show that in all settings, the proposed method manages to reduce the computational time significantly (up to 5 times faster) with a higher success rate. The method is very general and easily applicable to other robotic systems.



\section*{APPENDIX}
\subsection{Cost Functions for Projection and IK}
\label{sec:cost_functions}
We list here the cost functions for the projection and IK. Each cost function is a square function of the corresponding residual. We use the fast rigid body dynamics library \emph{pinocchio}~\cite{pinocchioweb} to compute the forward kinematics and the cost derivatives analytically. The optimization is stopped when each residual's norm is lower than the specified threshold.

\subsubsection{End-effector pose cost}
The residual for this cost is defined as
\begin{equation}
r_\text{ee}(\bm{q}) = \text{log}(\bm{T}_{\text{ref}} ^{-1} \bm{T}(\bm{q})),
    \label{eq:cose_pose}
\end{equation}
where $\bm{T}_{\text{ref}}$ and $\bm{T}(\bm{q})$ are the reference and the current pose, represented as SE(3), and log is the logarithm function that map SE(3) to se(3). The residual has 6 components, 3 for the position and 3 for the orientation. Depending on the constraints, we can set different weights for each of these components. For example, to set the orientation constraint for the Panda robot such that the gripper is always horizontal, the weights for the last rotation component and for the position components are set to zero, while the rest is set to one.

\subsubsection{Posture cost}
\begin{equation}
\bm{r}_p(\bm{q}) = (\bm{q} - \bm{q}_\text{nom}).
\end{equation}
$\bm{r}_p(\bm{q})$ measures the distance of the configuration $\bm{q}$ from a nominal configuration $\bm{q}_\text{nom}$. In this paper we often use the initial values given to the projector as $\bm{q}_\text{nom}$, so that the projector will project the configuration to the manifold while keeping it as close as possible to the initial configuration. This is especially important for the  \emph{ConstrainedExtend} step, because without this, the interpolated configurations may be projected differently, causing them to be discontinuous.

\subsubsection{Joint limit cost}
Given the joint lower and upper limit $(\bm{l}_{l}, \bm{l}_{u})$, the residual cost associated to the joint limit is computed as the distance to the nearest limit when the corresponding joint angle is out of the joint limits,
\begin{equation}
c_l(\bm{q}) = \min(\bm{q} - \bm{l}_{l}, \bm{0}) + \max(\bm{q} - \bm{l}_{u}, \bm{0}).
\end{equation}
The function $\min$ and $\max$ set the residual component to zero if the corresponding joint angle is within the bounds.

\subsubsection{Static stability cost}
To achieve static stability, the center of mass (CoM) projection on the ground should fall on the support polygon formed by the feet. We approximate this polygon by a rectangle surrounding the feet. The lower and upper limit of the horizontal CoM values, $(\bm{l}_{l}, \bm{l}_{u})$, are given by this rectangle.
Similar to the joint limit cost, the cost residual for the static stability is given as
\begin{equation}
\bm{r}_s(\bm{q}) = \min(\bm{p}_{\text{com}} - \bm{l}_{l}, \bm{0}) + \max(\bm{p}_{\text{com}} - \bm{l}_{u}, \bm{0}),
\end{equation}
where $\bm{p}_\text{com}$ is the center of mass position at $\bm{q}$.

%
%
%
%
%

\bibliographystyle{IEEEtran}
\bibliography{references}  

\end{document}